\documentclass{article}
\usepackage{times}
\usepackage{epsfig}
\usepackage{graphicx}
\usepackage{amsmath}
\usepackage{amssymb}
\usepackage{algorithm2e}
\usepackage{comment}
\usepackage{arxiv}

\usepackage[utf8]{inputenc} 
\usepackage[T1]{fontenc}    
\usepackage{hyperref}       
\usepackage{url}            
\usepackage{booktabs}       
\usepackage{amsfonts}       
\usepackage{nicefrac}       
\usepackage{microtype}      
\usepackage{cleveref}       
\usepackage{lipsum}         
\usepackage{graphicx}
\usepackage{natbib}
\usepackage{doi}

\title{A template for the \emph{arxiv} style}

\author{ \href{https://orcid.org/0000-0000-0000-0000}{\includegraphics[scale=0.06]{orcid.pdf}\hspace{1mm}David S.~Hippocampus}\thanks{Use footnote for providing further
		information about author (webpage, alternative
		address)---\emph{not} for acknowledging funding agencies.} \\
	Department of Computer Science\\
	Cranberry-Lemon University\\
	Pittsburgh, PA 15213 \\
	\texttt{hippo@cs.cranberry-lemon.edu} \\
	\And
	\href{https://orcid.org/0000-0000-0000-0000}{\includegraphics[scale=0.06]{orcid.pdf}\hspace{1mm}Elias D.~Striatum} \\
	Department of Electrical Engineering\\
	Mount-Sheikh University\\
	Santa Narimana, Levand \\
	\texttt{stariate@ee.mount-sheikh.edu} \\
}

\renewcommand{\shorttitle}{\textit{arXiv} Template}

\hypersetup{
pdftitle={A template for the arxiv style},
pdfsubject={q-bio.NC, q-bio.QM},
pdfauthor={David S.~Hippocampus, Elias D.~Striatum},
pdfkeywords={First keyword, Second keyword, More},
}

\begin{document}

\title{Generative Target Update for Adaptive\\ Siamese Tracking}

\author{Madhu Kiran\thanks{Laboratoire d’imagerie, de vision et d’intelligence artificielle (LIVIA), Ecole de technologie
superieure, Montreal, Canada}\thanks{Corresponding author, madhu\_sajc@hotmail.com} \and Le Thanh Nguyen-Meidine\footnotemark[1] \and Rajat Sahay\thanks{Vellore Institute of Technology, Vellore} \and Rafael Menelau Oliveira E Cruz\footnotemark[1] \and Louis-Antoine Blais-Morin\thanks{Genetec Inc.} \and Eric Granger\footnotemark[1]}

\maketitle

\begin{abstract}
  Siamese trackers perform similarity matching with templates (i.e., target models) to recursively localize objects within a search region. Several strategies have been proposed in the literature to update a template based on the tracker output, typically extracted from the target search region in the current frame, and thereby mitigate the effects of target drift. However, this may lead to corrupted templates, limiting the potential benefits of a template update strategy. 
   This paper proposes a model adaptation method for Siamese trackers that uses a generative model to produce a synthetic template from the object search regions of several previous frames, rather than directly using the tracker output. Since the search region encompasses the target, attention from the search region is used for robust model adaptation. In particular, our approach relies on an auto-encoder trained through adversarial learning to detect changes in a target object's appearance, and predict a future target template, using a set of target templates localized from tracker outputs at previous frames. To prevent template corruption during the update, the proposed tracker also performs change detection using the generative model to suspend updates until the tracker stabilizes, and robust matching can resume through dynamic template fusion.
   Extensive experiments conducted on  VOT-16, VOT-17, OTB-50, and OTB-100 datasets highlight the effectiveness of our method, along with the impact of its key components. Results indicate that our proposed approach can outperform state-of-art trackers, and its overall robustness allows tracking for a longer time before failure.\\
   \textbf{Code:} \url{https://anonymous.4open.science/r/AdaptiveSiamese-CE78/}
\end{abstract}

\section{Introduction}

Many video analytics, monitoring, and surveillance applications rely on visual object tracking (VOT) to locate targets appearing in a camera viewpoint over time, scene understanding, action and event recognition, video summarizing, person re-identification. In real-world video surveillance applications, VOT is challenging due to real-time computational constraints, changes and deformation in target appearance, rapid motions, occlusion, motion blur, and complex backgrounds. In real-time  video surveillance applications, the time required to capture and identify various events is a significant constraint. 
\begin{figure}[h!]
 \centering
 \includegraphics[width=0.7\linewidth]{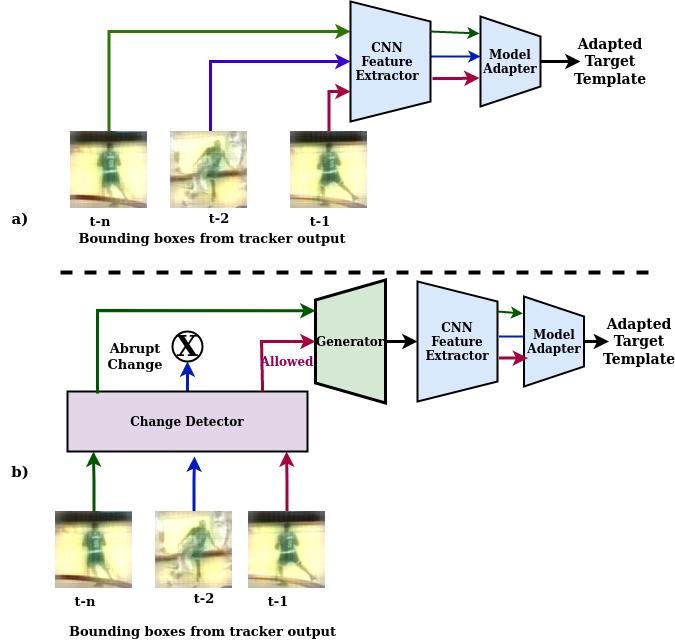}
   \caption{Approaches to select templates for adaptive Siamese tracking. (a) Conventional approaches select templates from previous tracker outputs. (b) Our approach generates templates from previous ones using a generative model, and filters noisy templates via change detection.}
  \label{fig:method} 
\end{figure}
Techniques for VOT may be categorized according to the target model or template construction mechanism, as either generative or discriminative. Generative appearance models represent target appearance without considering the background, while discriminative trackers learn a representation to distinguish between a target and background~\cite{salti2012}. The trackers can be further classified based on their image representation techniques, ranging from conventional hand-crafted descriptors~\cite{struck,6870486,cmt,kcf} to more recent deep learning models, like Siamese trackers~\cite{bertinetto2016fully,siamcar,SiamRPN,SiamVGG,SiamDW2,oceantracking,DaSiam}. 


One of the initial Siamese trackers -- the Fully Convolutional Siamese tracker (SiameseFC)~\cite{bertinetto2016fully} -- uses a single features representation extracted at the beginning of a trajectory, and does not update the target features during tracking. Although this strategy can provide computational efficiency, SiameseFC trackers suffer from target drift over time. Target drift is defined as a situation when the tracker slowly starts focusing on distractive backgrounds (rather than the target), and eventually looses the target. Such drift causes broken tracklets, a potential problem in video surveillance applications such as loitering detection, video person re-identification, face recognition, and other related applications. When the object's appearance changes abruptly, or the object is occluded or partially leaves the search region, the SiameseFC tracker temporarily drifts to a location with a high response map score~\cite{updtnet}. Some adaptive Siamese trackers have been proposed that allow for template updates. Most early trackers sought to update target features as a moving average based on localizations from the current frame output. Other trackers apply strategies to address drifting objects by storing an array of templates, or combining features from various tracker outputs~\cite{duynamicmem,updtnet}. However, these trackers face issues when updating templates on every frame based on tracker output. In particular, they integrate noise from the tracker output templates, especially in the present image occlusion or drift. Moreover, when training a Siamese tracker for matching based on multiple templates, learning the template update function in addition to the conventional search-template pair may lead to over-fitting~\cite{updtnet}. Hence, to avoid corrupting templates, it is important to manage when and how the templates are  updated.

In this paper, we focus robust VOT of a single object, where the template is updated dynamically in response to changes in the object's appearance. 
This paper introduces a method that applied to any adaptive Siamese trackers for real-time applications. Instead of using the samples mined directly from the tracker output, we propose to use a generative model to generate a sample observing many previous target template. This generative model predicts the future appearance of a target template given a set of consecutive target templates localized from tracker outputs at previous frames. It also allows detecting abrupt changes in the appearance of target objects, and thereby preventing template corruption by suspending template updates until the tracker stabilizes. In the absence of an abrupt change, our generative model outputs a synthetic target template for robust matching through dynamic template fusion, and updating the target template. 

In contrast with~\cite{updtnet}, our method learns the target update itself, using cross-attention between search region and template features. This allows selecting channels among the target features that are most useful for target update. The cross-attention approach relies on attention from the target's current appearance in the search region to update the existing target template. The proposed generative model is designed by adversarial training a video autoencoder to produce a future frame. The discrepancy between the generated future frame, and the target's appearance from tracker output helps detect appearance changes using a change detection mechanism. We summarise our contribution as follows. We propose a method for adaptation of Siamese trackers based generative model update. The generative model produces a future template by observing the the past templates. Additionally, change detection is proposed using the generative model to suspend model update during target drifting. Finally, the method relies on the difference between a simple average and a learned fusion templates to define an inequality constraint during learning of model adaptation. It uses attention from the search region to attend to salient regions in the tracker localised template. For proof-of-concept validation, the proposed method is integrated into state-of-art SiamFC+ and SiamRPN trackers~\cite{SiamDW2,SiamRPN}, and compared to different conventional and state-of-art trackers from deep Siamese family~\cite{bertinetto2016fully,SiamDW2} on videos from the OTB~\cite{otb} and VOT~\cite{vot2017,vot2018} evaluations datasets. We also perform ablation studies on different modules to study the effectiveness of the proposed method.

\section{Related Work}

Pioneered by SINT~\cite{sint} and SiamFC~\cite{bertinetto2016fully}, the Siamese family of trackers evolved from Siamese networks trained offline with similarity metrics. These networks were trained on a large dataset to learn generic features for object tracking. SiamRPN~\cite{SiamRPN} further improves on this work by employing region proposals to produce a target-specific anchor-based detector. Then, the following Siamese trackers mainly involved designing more powerful backbones~\cite{SiamDW2,SiamVGG} or proposal networks, like in~\cite{cascaded}. ATOM~\cite{atom} and DIMP~\cite{dimp} are robust online trackers that differ from the general offline Siamese trackers by their ability to update model online during tracking. Other paradigms of Siamese trackers are distractor-aware training, domain-specific tracking~\cite{SA-Siam,DaSiam}.

In \cite{reinforcement}, an LSTM is incorporated to learn long-term relationships during tracking and turns the VOT problem into a consecutive decision-making process of selecting the best model for tracking via reinforcement learning~\cite{regularity1}. In~\cite{correl} and~\cite{DaSiam}, models are updated online by a moving average based learning. These methods integrate the target region extracted from tracker output into the initial target. In~\cite{generative1}, a generative model is learned via adversarial learning to generate random masks that produce shifted versions of target templates from the original template. Then, an adversarial function is used to decide whether or not the generated template is from the same distribution and if they will be used as templates for tracking. In~\cite{duynamicmem}, an LSTM is employed to estimate the current template by storing previous templates in a  memory bank. In~\cite{dynamic}, authors propose to compute transformation matrix with reference to the initial template, with a regularised linear regression in the Fourier domain. Finally, in~\cite{yao2018joint}, authors propose to learn the updating co-efficient of a correlation filter-based tracker using SGD online. All these methods use the tracker output as the reference template while updating on top of the initial template.~\cite{dimp,prdimp} propose a model where an adaptive discriminative model is generated online by the steepest gradient descent method. They differ from another online learned method like ~\cite{nam2016mdnet} due to their real-time performance. Similarly~\cite{oceantracking} introduce online model prediction but employ a fast conjugate gradient algorithm for model prediction. Foreground score maps are estimated online, and the classification branch is combined by weighted addition. 

Several methods follow the standard strategy of updating target template features, such as simple averaging, where the template is updated as a running average with exponentially decaying weights over time. This yields a template update defined by: 
\begin{equation}
\label{eqn:exponential}
\widetilde{\varphi}^{n}=(1-\gamma) \widetilde{\varphi}^{n-1}+\gamma \varphi^{n} , 
\end{equation}
where $n$ denotes the time step, $\widetilde{\varphi}^{n}$ the predicted template, and $\gamma$ the learning rate.

This strategy has several issues, most notably the possibility of integrating noise and corruption into templates during the update process. Therefore, authors in~\cite{updtnet}  proposed a network which, when given an input of past template features, the template extracted from current tracker output produces a new representation that can be added to the original ground truth template (obtained during tracker initialization). This approach further suffers from the following issues. (1) A future template for the tracker is unseen at the time of template update, and the model is updated solely based on the tracker output in the past frame output. (2) The model is updated every frame making it still susceptible to the integration of noise over time. (3) Network training is a tedious task since it must be trained continuously offline by running the tracker on the training dataset. It must produce a previous frame feature representation that needs to be stored and used for the next training iteration. Further developments in this direction are challenging.


\section{Proposed Adaptive Siamese Tracker}

Given the initial object location, a ground truth-object template image $T$ is extracted, along with the corresponding deep CNN features $\varphi_{gt}$. A tracker seeks to produce object localization $BBox$ at a given time step by matching $\varphi_{gt}$ with search region features $\varphi_s$. The objective is to produce a trajectory by recursively extracting search regions from tracker output, and matching them with a given template over each input video frame. 

\paragraph{\textbf{a) Template Prediction and Change Detection:}}
Inspired from ~\cite{tang2020integrating}, We employ a video autoencoder that is trained through adversarial learning for template generation. Given an set of past templates $T^n$ where $n = t, t-1, t-2, t-3...$ we aim to predict a future template for time step $t$. As described below, our template generation method consists of a generator and a discriminator.
%
%
%
\begin{figure}[b!]
 \centering
\includegraphics[width=1.0\linewidth]{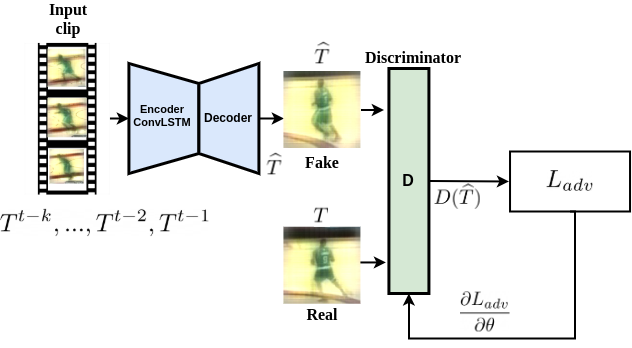}
   \caption{Our generator model is a video autoencoder that is trained adversarially. A future target template is reconstructed from a sequence of input target templates. The discriminator $D$ processed the reconstructed template as fake, and the ground truth template input as real.}
  \label{fig:gan} 
\end{figure}

\paragraph{Generator:}
It consists of an encoder-decoder architecture (see Fig~\ref{fig:gan}). The encoder compresses an input video clip into a small bottleneck with a set of CNN layers and Conv-LSTM based recurrent network to model the temporal relationship between the frames. The decoder consists of some layers of transposed CNN to obtain the predicted video frame. Hence given an input video clip of $T^{t-k},...,T^{t-2},T^{t-1}$, the Generator produces the estimated future video frame $\hat{T}$. The generator is trained according to the Mean Squared Error (MSE) between predicted image $\widehat{T}$ and ground truth image  $T$
\begin{figure*}[t]
 \centering
\includegraphics[width=1.0\linewidth]{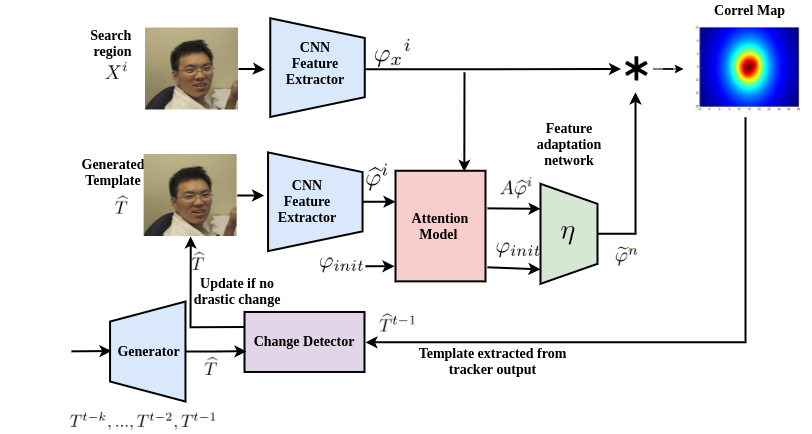}
   \caption{Block diagram of our proposed generic template update system for Siamese trackers that adapts the model of the target template with a generative model and change detection. Our attention based dynamic ensemble of targets adapts the model to the current representation of the target with attention from search region. The change detection system disables template update during anomalies such as occlusion and severe target drift.}
  \label{fig:overall} 
\end{figure*}

\paragraph{Discriminator:}
It comprises of several CNN layers to compete with the generator to differentiate between the ground truth and generated frames. The discriminator distinguishes a real-world image from a fake image, promoting the generator to produce good quality images. Since training the autoencoder on MSE loss alone will cause the output to be blurry, we leverage the discriminator to help produce higher-quality images. The labels are set to 0 for fake images (obtained from the autoencoder's reconstruction) and 1 for real (ground truth template image). The discriminator is trained with an adversarial loss:
\begin{equation}
\label{eqn:adversarial}
L_{a d v}^{D}(T, \widehat{T})=\frac{1}{2}(D(T)-1)^{2}+\frac{1}{2}(D(\widehat{T})-0)^{2}
\end{equation}

\paragraph{Change Detection:}
Once the adversarial auto-encoder has been trained, the average MSE error between the reconstructed template and search regions from each input frame in the video clip is computed to produce the reconstruction error. Similar to previous methods such as~\cite{regularity1,videoae}, we adopt the regularity score to detect abrupt changes in template clips. Let $e(T)$ be the reconstruction error. Reconstruction error should be normalized from the sequences of the same video with: 
\begin{equation}
\label{eqn:regularity}
s(x)=1-\frac{e(T)-\min _{T} e(T)}{\max _{T} e(T)}
\end{equation}

In practice it is difficult to set $\min _{T} e(T)$ and $\max _{T} e(T)$ as the future frames are not observable. Hence, we set $\min _{T} e(T)$ and $\max _{T} e(T)$ experimentally using a validation set. The regularity score $s(x)$ serves as the measure for regular templates. Hence a score of less than a threshold $\tau$ is considered an abrupt change. The length of the input template sequence is kept fixed, and new templates are updated into the sequence by pushing the oldest template out of the stack. When a change is detected, the template that was last pushed into the stack is rejected and considered a possible source of corruption, and the template update eventually stalled for that particular time step.

\paragraph{\textbf{b) Template Update with Cross Attention:}}
Target model adaptation is often based on the last known appearance of the object from previous frames. At the start of the tracking, the initial target feature $\varphi_{init}$ needs to be adapted to match the latest object appearance. Such adaptation is not possible without predicting the tracker in the current frame. At the same time, it is to be noted that the search region encompasses the target in the current frame, given that the change detector has detected no drastic change. Therefore we propose to use this cue to obtain attention from the search region to adapt the model. In addition to this, search region features and template features are of different sizes. This difference in feature size has inspired our proposal of using channel attention across the search and template stream. We follow a similar model adaptation paradigm as ~\cite{updtnet} along with our attention model and proposed optimization with inequality constraints.\cite{updtnet} consider adapting the target feature by adding additional information to the initial target feature $\varphi_{init}$.

Let $\widehat{\varphi}^t$ be the feature extracted from the generated template $\widehat{T}^t$. The generated template is the predicted target appearance. In comparison to $\widehat{\varphi}^t$,  $\varphi_{init}$ is the most reliable target feature as it is obtained during the initialization of the tracker using ground truth information. The model adaptation mechanism considers both $\varphi_{init}$ and $\widehat{\varphi}^t$ to predict the adapted feature $\widetilde{\varphi}^{t}$. As discussed earlier, the first step is to obtain attention from the search region to select important channels in $\widehat{\varphi}^t$.
Let ${\varphi_z}^{t}$ be the feature extracted from search region. Then we obtain matching channel attention from ${\varphi_z}^{t}$, $\varphi_{init}$ and $\widehat{\varphi}^t$ by passing through an attention model similar to channel attention in ~\cite{cosam} , using an MLP with Sigmoid activation to select channels based on importance. The attention obtained from ${\varphi_z}^{t}$, $\varphi_{init}$ and $\widehat{\varphi}^t$ are averaged to obtain channel attention $A$.  Attention $A$ is multiplied with $\widehat{\varphi}^t$. Therefore the channels of $\widehat{\varphi}^t$ have been re-weighed and common saliency across search region and target template are encompassed into the attention. 

The attended feature $A \widehat{\varphi}^t$ and $\widetilde{\varphi}^{t-1}$ (obtained from the prior frame after model adaptation) are then concatenating in the channel dimension as follows~\ref{eqn:concat}:
\begin{equation}
\label{eqn:concat}
 \varphi_{concat}= [\widetilde{\varphi}^{t-1} ;A \widehat{\varphi}^t ]
\end{equation}
The concatenated feature $\varphi_{concat}$ is passed through a two layer CNN with 1x1 convolution layer, followed by a TanH activation function to obtain adapted feature in: 
\begin{equation}
\label{eqn:adapted}
\widetilde{\varphi}^{t}= \mathbf{\eta}(\varphi_{concat}) , 
\end{equation}
where $\mathbf{\eta}$ is the model adaptation network discussed above, and $\widetilde{\varphi}^{t}$ is the adapted target template for tracking.

\paragraph{\textbf{c) Model Adaptation:}}
During training,  target samples are generated from the training data keeping the chronological order of the image frame in a video to obtain features $\varphi_{init}, \varphi_{GT} $. The ground truth video data generated these two, i.e., initial and template from future frames. To obtain the generated template, $n$ consecutive templates are used from the same video to generate $\widehat{\varphi}^t$ by using the pre-trained generator that was previously discussed. To enable the system, learn to generate an adapted feature to resemble a target template from the next frame, we employ MSE loss:
\begin{equation}
\label{eqn:mse_final}
 L_{mdl-mse}=\|\varphi_{GT}-\widetilde{\varphi}^{t}\|_{2}, 
\end{equation}
where $\varphi_{GT}$ are the ground truth target features which are chronologically the latest template. We expect the adapted template $\widetilde{\varphi}^{t}$  obtained by adapting previously seen target templates to resemble the future ground truth template.

Optimizing the MSE loss in our case is a difficult task since the model is being forced to learn to produce an unseen representation from future frames given two different previously seen frames. In~\cite{updtnet}, the tracker is recursively train on several training cycles, which is a tedious task. Template update can also be performed by simply averaging features that would suffer from noisy updates and feature smoothing due to averaging both leading to information loss. Such simple averaging can be used as a cue to introduce a constraint to optimize the template update. 

Let $\varphi_{avg}$ be the averaged template obtained by averaging $\varphi_{init}$ and $\varphi^{t-1}$. Let $D_{E}$ denote the Euclidean distance function. It is reasonable to assume that simple template averaging is a trivial solution and therefore the distance between learnt template $\widetilde{\varphi}^{t}$ and $\varphi_{GT}$(the future template)  must be less than $\varphi_{avg}$ and $\varphi_{GT}$. Constrained loss given by, 
\begin{equation}
\label{eqn:mse_optim}
 \hfill L_{const-mse}=\|\varphi_{GT}-\widetilde{\varphi}^{t}\|_{2} + \\ \lambda~ ReLU( (D_{E}(\varphi_{GT}, \widetilde{\varphi}) - D_{E}(\varphi_{GT}, \varphi_{avg}) )
\end{equation}
where ReLU ensures that the gradients are passed for the constraint only when the constraint is not respected. $\lambda$ is set to a value $\gg 1$ and is determined experimentally.

\section{Results and Discussion}

\paragraph{\textbf{a) Experimental Methodology:}}
A ResNet-22 CNN similar to SiamDW tracker~\cite{smeulders,SiamDW2} is used for a fair comparison. The system on GOT-10K dataset~\cite{got10k} to train our video autoencoder, as well as the tracking network similar to~\cite{smeulders} for direct comparison since they use a similar baseline as ours. GOT-10K has around 10,000 video sequences with 1.5 million labeled bounding boxes to train our tracking network and auto encoder. In particular, due to many training sequences, the autoencoder overall motion model for objects in generic videos to predict frames in the future. We used the official training set of GOT10-K to train the networks. We use the same data augmentation techniques as ~\cite{smeulders,SiamDW2}. The autoencoder was pre-trained adversarially with the discriminator. The Siamese tracker is pre-trained without the autoencoder by selecting random samples in a specific video, one for the template and the other for the search region. 

The standard tracking benchmarks, OTB2013, OTB2015~\cite{otb} and VOT2017~\cite{vot2017} video datasets, are uses to evaluate trackers. The OTB~\cite{otb} dataset consists of sets OTB213 and OTB2015 with 50 and 100 real-world tracking videos, respectively. The metrics used with OTB datasets are success rate and precision. VOT2017 dataset has 60 public test videos with a total of 21,356 frames. The VOT protocol re-initializes the tracker when the tracker fails with a delay of 5 frames. Evaluation measures used with VOT are EAO and (Expected average overlap), a combination of accuracy and robustness. Robustness refers to the number of times a tracker needs to be initialized.

\begin{table*}[t!]
\caption{EAO and robustness associated with different components of our proposed tracker on the VOT2017 dataset.}
\begin{center}
{
\begin{tabular}{|l|l|l|cc|}
\hline
\textbf{Sl} & \textbf{Ablation}                & \textbf{Remark}                     & \textbf{EAO$\boldsymbol{\uparrow}$}  & \textbf{Robustness$\boldsymbol{\downarrow}$}     \\ \hline\hline
\multicolumn{5}{|l|}{\textbf{$\cdot$ Template update}}                                          \\ \hline
1  & Only SiamFC+            & Baseline                    & 0.23 & 0.49                \\
2  & SiamFC+ and UpdateNet      & Baseline and  Update              & 0.26 & 0.40       \\
3  & SiamFC+ and Moving Average  & Baseline and  Linear           & 0.25 & 0.44         \\
4  & SiamFC+ and Dynamic Update & Ours without Constraint     & 0.27 & 0.41             \\
5  & SiamFC+ and Dynamic Constr    & Ours with INQ. Constraint   & 0.29 & 0.38            \\ \hline
\multicolumn{5}{|l|}{\textbf{$\cdot$ Generative Modelling}}                                         \\ \hline
6  & Generated Template Update  & 5) + Generated Template     & 0.29 & 0.37             \\
7  & Generated Model and  Blend   & 6) + Tracker Output Blend  & 0.30  & 0.37           \\ \hline
\multicolumn{5}{|l|}{\textbf{$\cdot$ Change Detection}}                                         \\ \hline
8  & Change Detection   & 7) + No Update on Drastic Change & 0.31 & 0.34                \\ \hline
\end{tabular}
}
\end{center}
\label{tab:abla}
\end{table*}

\paragraph{\textbf{b) Ablation Study:}}
We study the contribution of different components of our proposed method on the VOT2017 dataset. In the first part of Tab~\ref{tab:abla}, "Template update," demonstrates our contribution to model adaptation. The second part, "Generative Model," evaluates the contribution of the generative model in the template update. Finally, the "Change Detection" part shows the effect of change detection on tracking EAO.
%
In order to evaluate the template update part, we compare the results of the baseline ~\cite{SiamDW2} which is also our backbone. The template update mechanism uses the output from tracker instead of the generative model instead of $\widehat{\varphi}^t$ in the template update network. We implement ~\cite{updtnet} based model adaptation for the baseline ~\cite{SiamDW2} and moving average based linear update as in ~\cite{DaSiam} is compared with our proposed update method "Dynamic Update" (with attention), which refers to training without the inequality constraint discussed above. Number 5) in the table refers to the experiment where template update is used with inequality constraints. It can be seen that using the inequality constraint alone and our template update mechanism has improved the overall Robustness of the tracker as indicated by the robustness score(lower the score more robust the tracker is). 6) and 7) in the Tab.~\ref{tab:abla} uses the output from generative model to feed $\widehat{\varphi}^t$. Since the generative model's output is a bit blurry in 7) we blend it with tracker output extracted target template image to obtain a sharper image. Such blending has been shown to improve the result further. We detect drastic changes in the model via the regularity score of the tracker. The change detection will help prevent noisy updates during drift or occlusion; this is shown in 8) where no updates were made during drastic changes. 

\begin{table*}[!t]
\caption{Accuracy of our proposed and state-of-art trackers on the OTB-50, OTB-100, VOT2016 and VOT2017 datasets.}
\begin{center}
\scalebox{0.8}
{
\begin{tabular}{|l||lllllllllllll|}
\hline
\textbf{Tracker} & \multicolumn{2}{l}{\textbf{OTB2013}} &  & \multicolumn{2}{l}{\textbf{OTB2015}} &  & \multicolumn{3}{c}{\textbf{VOT2016}} &  & \multicolumn{3}{c|}{\textbf{VOT2017}} \\ \cline{2-3} \cline{5-6} \cline{8-10} \cline{12-14} 
 & AUC$\boldsymbol{\uparrow}$ & Prec$\boldsymbol{\uparrow}$ &  & AUC$\boldsymbol{\uparrow}$ & Prec$\boldsymbol{\uparrow}$ &  & EAO$\boldsymbol{\uparrow}$ & A$\boldsymbol{\uparrow}$ & R$\boldsymbol{\downarrow}$ &  & EAO$\boldsymbol{\uparrow}$ & A$\boldsymbol{\uparrow}$ & R$\boldsymbol{\downarrow}$ \\ \cline{1-3} \cline{5-6} \cline{8-10} \cline{12-14}  \hline \hline
SINT, CVPR-16~\cite{sint} & 0.64 & 0.85 &  & - & - &  & - &  & - &  & - &  & - \\
SiamFC, ECCV-16~\cite{bertinetto2016fully} & 0.61 & 0.81 &  & 0.58 & 0.77 &  & 0.24 & 0.53 & 0.46 &  & 0.19 & 0.5 & 0.59 \\
DSiam, ECCV-17~\cite{DaSiam} & 0.64 & 0.81 &  & 0.64 & 0.81 &  & - &  & - &  & - &  & - \\
StructSiam, ECCV-18~\cite{structsiam} & 0.64 & 0.88 &  & 0.62 & 0.85 &  &  & - & - &  & - & - & - \\
TriSiam, ECCV-18,~\cite{siamtriplet} & 0.62 & 0.82 &  & 0.59 & 0.78 &  & - & - & - &  & 0.2 &  & - \\
SiamRPN, CVPR-18~\cite{SiamRPN} & - & - &  & 0.64 & 0.85 &  & 0.34 & 0.56 & 0.26 &  & 0.24 & 0.49 & 0.46 \\

SE-Siam, WACV-21~\cite{smeulders} & 0.68 & 0.90 &  & 0.66 & 0.88 &  & 0.36 & 0.59 & 0.24 &  & 0.27 & 0.54 & 0.38 \\ 
SiamFC+, CVPR-19~\cite{SiamDW2} & 0.67 & 0.88 &  & - & - &  & 0.30 & 0.54 & 0.26 &  & 0.24 & 0.49 & 0.46 \\

SiamRPN++, CVPR-19~\cite{siamrpnPP} & - &- &   & 0.69 & 0.89 &  & 0.46 &0.64& 0.20 &  & 0.41 & 0.60 & 0.23 \\ \hline \hline
Adaptive SiamFC+ (ours) & 0.68 & 0.89 &   & 0.67& \textbf{0.89} &  & \textbf{0.39} &0.56& 0.21 &  & \textbf{0.31} & 0.52 & \textbf{0.34} \\  
Adaptive SiamRPN++ (ours) & - &- &   & \textbf{0.71} & 0.87 &  & 0.47 &0.61& \textbf{0.19} &  & \textbf{0.44} & 0.58 & \textbf{0.21} \\ \hline

\end{tabular}
}
\end{center}
\label{tab:soa}
\end{table*}

 \paragraph{\textbf{c) Comparison with State-of-Art:}}
 We compare our proposed template update method implemented on SiamFC+~\cite{SiamDW2} back-end against popular Siamese methods bench marked on OTB-50,OTB-100,VOT16,17 datasets. Similar to the benchmarking method in SE-SiamFC~\cite{smeulders} we have selected the Siamese trackers for direct comparison with ours. It is important to note that our back-end Siamese tracker, training procedure, sample selection, Etc., are the same as~\cite{smeulders}. 
 OTB benchmark uses AUC, which signifies the average overlap over the dataset, and Precision (Prec) signifies the center distance error between object and tracker bounding box. We can see that our method performs competitively with ~\cite{smeulders} on OTB dataset shown in Tab.\ref{tab:soa} . It is important to note that OTB does not re-initialize the tracker on failure, and in addition, OTB does not consider track failures into the final evaluation.
 
 On the other hand, the VOT dataset uses Expected average Overlap (EAO), Robustness (R), and Accuracy (A) as metrics. Particularly Robustness is interesting as it indicates some measure on tracker drift in a given dataset. EAO combines tracking accuracy and Robustness, and hence it is a better indicator of tracker performance than just AUC. We can see from the Tab.\ref{tab:soa} our method outperforms SOA by 4\% and outperforms the baseline SiamFC+~\cite{SiamDW2} by 7\% on EAO. The results show that our proposed method would enable the tracker to track for longer periods before complete failure compared to the other methods we compare.

\begin{figure*}[h!]
 \centering
 \includegraphics[width=0.99\linewidth]{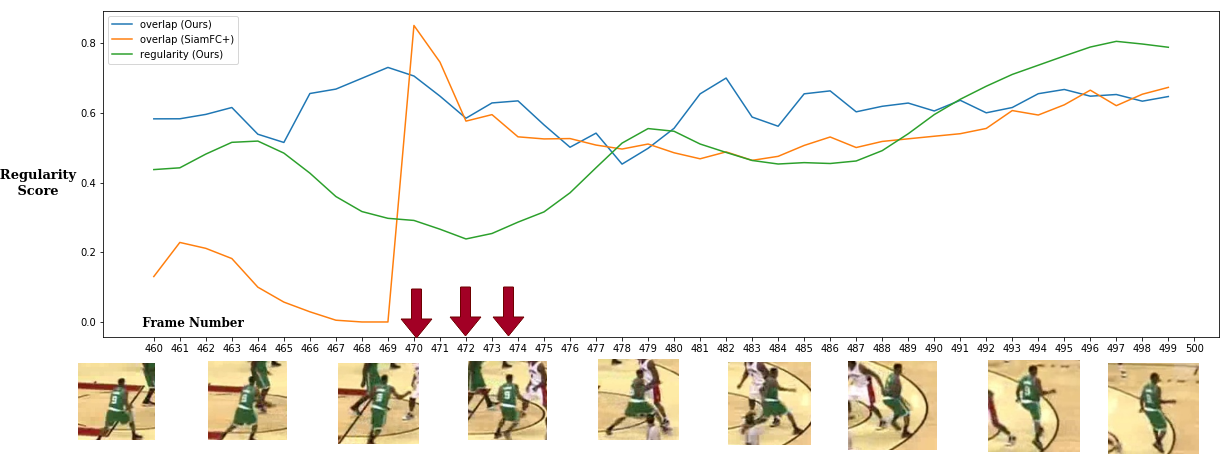}
   \caption{Visualization of tracker accuracy in terms of instantaneous overlap (overlap) of tracker output with ground truth bounding box with video frame number on x axis. We show the results for the trackers with our proposed model update and the baseline SiamFC. Red arrow on the x-axis indicates points of drastic changes.}
  \label{fig:reg} 
\end{figure*}

To show drastic changes during tracking, we plot the IOU "overlap" (intersection over union for tracking bounding box over ground truth) and the regularity score produced by our change detector. In Fig~\ref{fig:reg} blue line indicates IOU for our proposed tracker. The thumbnails at the bottom indicate cutouts of the ground truth bounding box around the object being tracked. The video example is from "basketball" of the VOT17 dataset. It can be observed that the regularity score produced by our change detector is low during frames that have partial occlusion and during clutter around the background. 

 \section{Conclusion}
 
 Adaptive Siamese trackers commonly rely on the tracker's output to update the target model. In this paper, we have identified shortcomings with this approach, and proposed a generative model to predict a synthetic target template based on the appearance of several templates from previous time steps. Since the generative model learns the future template from the distribution over past time steps, it suppresses stochastic noise. We also propose a change detection mechanism to avoid noisy updates during abrupt changes in target appearance. Our proposed method can be integrated into any Siamese tracker, and results achieved on VOT16, VOT17, OTB-50, and OTB-100 datasets indicate that it can provide a high level of robustness (can track for a longer period before drifting) compared to state-of-art adaptive and baseline trackers.

\bibliographystyle{unsrtnat}
\bibliography{egbib}

\begin{thebibliography}{38}
\providecommand{\natexlab}[1]{#1}
\providecommand{\url}[1]{\texttt{#1}}
\expandafter\ifx\csname urlstyle\endcsname\relax
  \providecommand{\doi}[1]{doi: #1}\else
  \providecommand{\doi}{doi: \begingroup \urlstyle{rm}\Url}\fi

\bibitem[Salti et~al.(2012)Salti, Cavallaro, and Stefano]{salti2012}
S.~Salti, A.~Cavallaro, and L.~Di Stefano.
\newblock Adaptive appearance modeling for video tracking: Survey and
  evaluation.
\newblock \emph{IEEE Transactions on Image Processing}, 21\penalty0
  (10):\penalty0 4334--4348, 2012.

\bibitem[Hare et~al.(2016)Hare, Golodetz, Saffari, Vineet, Cheng, Hicks, and
  Torr]{struck}
S.~Hare, S.~Golodetz, A.~Saffari, V.~Vineet, M.~M. Cheng, S.~L. Hicks, and
  P.~H.~S. Torr.
\newblock Struck: Structured output tracking with kernels.
\newblock \emph{IEEE Trans. PAMI}, 38\penalty0 (10):\penalty0 2096--2109, 2016.

\bibitem[Henriques et~al.(2015)Henriques, Caseiro, Martins, and
  Batista]{6870486}
J.~F. Henriques, R.~Caseiro, P.~Martins, and J.~Batista.
\newblock High-speed tracking with kernelized correlation filters.
\newblock \emph{IEEE Transactions on Pattern Analysis and Machine
  Intelligence}, 37\penalty0 (3):\penalty0 583--596, 2015.

\bibitem[Nebehay and Pflugfelder()]{cmt}
G.~Nebehay and R.~Pflugfelder.
\newblock Consensus-based matching and tracking of keypoints for object
  tracking.
\newblock In \emph{WACV 2014}, March .
\newblock \doi{10.1109/WACV.2014.6836013}.

\bibitem[Wang et~al.()Wang, O'Brien, Xiang, Xu, and Najjaran]{kcf}
X.~Wang, M.~O'Brien, C.~Xiang, B.~Xu, and H.~Najjaran.
\newblock Real-time visual tracking via robust kernelized correlation filter.
\newblock In \emph{ICRA 2017}.

\bibitem[Bertinetto et~al.(2016)Bertinetto, Valmadre, Henriques, Vedaldi, and
  Torr]{bertinetto2016fully}
Luca Bertinetto, Jack Valmadre, Jo{\~a}o~F Henriques, Andrea Vedaldi, and
  Philip~HS Torr.
\newblock Fully-convolutional siamese networks for object tracking.
\newblock \emph{arXiv:1606.09549}, 2016.

\bibitem[Guo et~al.({\natexlab{a}})Guo, Wang, Cui, Wang, and Chen]{siamcar}
Dongyan Guo, Jun Wang, Ying Cui, Zhenhua Wang, and Shengyong Chen.
\newblock Siamcar: Siamese fully convolutional classification and regression
  for visual tracking.
\newblock In \emph{CVPR2020}, {\natexlab{a}}.

\bibitem[Li et~al.({\natexlab{a}})Li, Yan, Wu, Zhu, and Hu]{SiamRPN}
Bo~Li, Junjie Yan, Wei Wu, Zheng Zhu, and Xiaolin Hu.
\newblock High performance visual tracking with siamese region proposal
  network.
\newblock In \emph{CVPR 2018}, {\natexlab{a}}.

\bibitem[Li and Zhang(2019)]{SiamVGG}
Yuhong Li and Xiaofan Zhang.
\newblock Siamvgg: Visual tracking using deeper siamese networks.
\newblock 2019.

\bibitem[Zhang and Peng()]{SiamDW2}
Zhipeng Zhang and Houwen Peng.
\newblock Deeper and wider siamese networks for real-time visual tracking.
\newblock In \emph{CVPR 2019}.

\bibitem[Zhang et~al.({\natexlab{a}})Zhang, Peng, Fu, Li, and
  Hu]{oceantracking}
Zhipeng Zhang, Houwen Peng, Jianlong Fu, Bing Li, and Weiming Hu.
\newblock Ocean: Object-aware anchor-free tracking.
\newblock In \emph{ECCV 2020}, {\natexlab{a}}.

\bibitem[Zhu et~al.()Zhu, Wang, Bo, Wu, Yan, and Hu]{DaSiam}
Zheng Zhu, Qiang Wang, Li~Bo, Wei Wu, Junjie Yan, and Weiming Hu.
\newblock Distractor-aware siamese networks for visual object tracking.
\newblock In \emph{ECCV 2018}.

\bibitem[Zhang et~al.({\natexlab{b}})Zhang, Gonzalez-Garcia, Weijer, Danelljan,
  and Khan]{updtnet}
Lichao Zhang, Abel Gonzalez-Garcia, Joost van~de Weijer, Martin Danelljan, and
  Fahad~Shahbaz Khan.
\newblock Learning the model update for siamese trackers.
\newblock In \emph{ICCV 2019}, {\natexlab{b}}.

\bibitem[Yang and Chan()]{duynamicmem}
Tianyu Yang and Antoni~B Chan.
\newblock Learning dynamic memory networks for object tracking.
\newblock In \emph{ECCV 2018}.

\bibitem[Wu et~al.()Wu, Lim, and Yang]{otb}
Yi~Wu, Jongwoo Lim, and Ming-Hsuan Yang.
\newblock Online object tracking: A benchmark.
\newblock In \emph{CVPR 2013}.

\bibitem[{Kristan} and et~al.()]{vot2017}
M.~{Kristan} and et~al.
\newblock The visual object tracking vot2017 challenge results.
\newblock In \emph{ICCVW 2017}.

\bibitem[Kristan and et~al.(2018)]{vot2018}
Matej Kristan and et~al.
\newblock The sixth visual object tracking vot2018 challenge results, 2018.

\bibitem[Tao et~al.()Tao, Gavves, and Smeulders]{sint}
Ran Tao, Efstratios Gavves, and Arnold~WM Smeulders.
\newblock Siamese instance search for tracking.
\newblock In \emph{CVPR 2016}.

\bibitem[Fan and Ling()]{cascaded}
Heng Fan and Haibin Ling.
\newblock Siamese cascaded region proposal networks for real-time visual
  tracking.
\newblock In \emph{CVPR2019}.

\bibitem[Danelljan et~al.({\natexlab{a}})Danelljan, Bhat, Khan, and
  Felsberg]{atom}
Martin Danelljan, Goutam Bhat, Fahad~Shahbaz Khan, and Michael Felsberg.
\newblock Atom: Accurate tracking by overlap maximization.
\newblock In \emph{CVPR2019}, {\natexlab{a}}.

\bibitem[Bhat et~al.(2019)Bhat, Danelljan, Gool, and Timofte]{dimp}
Goutam Bhat, Martin Danelljan, Luc~Van Gool, and Radu Timofte.
\newblock Learning discriminative model prediction for tracking.
\newblock In \emph{Proceedings of the IEEE/CVF International Conference on
  Computer Vision}, pages 6182--6191, 2019.

\bibitem[He et~al.()He, Luo, Tian, and Zeng]{SA-Siam}
Anfeng He, Chong Luo, Xinmei Tian, and Wenjun Zeng.
\newblock A twofold siamese network for real-time object tracking.
\newblock In \emph{CVPR 2018}.

\bibitem[Zhong et~al.(2018)Zhong, Bai, Li, Zhang, and Fu]{reinforcement}
Bineng Zhong, Bing Bai, Jun Li, Yulun Zhang, and Yun Fu.
\newblock Hierarchical tracking by reinforcement learning-based searching and
  coarse-to-fine verifying.
\newblock \emph{IEEE Transactions on Image Processing}, 28\penalty0
  (5):\penalty0 2331--2341, 2018.

\bibitem[Duman and Erdem(2019)]{regularity1}
Elvan Duman and Osman~Ayhan Erdem.
\newblock Anomaly detection in videos using optical flow and convolutional
  autoencoder.
\newblock \emph{IEEE Access}, 7:\penalty0 183914--183923, 2019.

\bibitem[Valmadre et~al.()Valmadre, Bertinetto, Henriques, Vedaldi, and
  Torr]{correl}
Jack Valmadre, Luca Bertinetto, Joao Henriques, Andrea Vedaldi, and Philip~HS
  Torr.
\newblock End-to-end representation learning for correlation filter based
  tracking.
\newblock In \emph{CVPR 2017}.

\bibitem[Song et~al.()Song, Ma, Wu, Gong, Bao, Zuo, Shen, Lau, and
  Yang]{generative1}
Yibing Song, Chao Ma, Xiaohe Wu, Lijun Gong, Linchao Bao, Wangmeng Zuo, Chunhua
  Shen, Rynson~WH Lau, and Ming-Hsuan Yang.
\newblock Vital: Visual tracking via adversarial learning.
\newblock In \emph{CVPR 2018}.

\bibitem[Guo et~al.({\natexlab{b}})Guo, Feng, Zhou, Huang, Wan, and
  Wang]{dynamic}
Qing Guo, Wei Feng, Ce~Zhou, Rui Huang, Liang Wan, and Song Wang.
\newblock Learning dynamic siamese network for visual object tracking.
\newblock In \emph{ICCV2017}, {\natexlab{b}}.

\bibitem[Yao et~al.()Yao, Wu, Zhang, Shan, and Zuo]{yao2018joint}
Yingjie Yao, Xiaohe Wu, Lei Zhang, Shiguang Shan, and Wangmeng Zuo.
\newblock Joint representation and truncated inference learning for correlation
  filter based tracking.
\newblock In \emph{ECCV 2018}.

\bibitem[Danelljan et~al.({\natexlab{b}})Danelljan, Gool, and Timofte]{prdimp}
Martin Danelljan, Luc~Van Gool, and Radu Timofte.
\newblock Probabilistic regression for visual tracking.
\newblock In \emph{CVPR 2020}, {\natexlab{b}}.

\bibitem[Nam and Han()]{nam2016mdnet}
Hyeonseob Nam and Bohyung Han.
\newblock Learning multi-domain convolutional neural networks for visual
  tracking.
\newblock In \emph{CVPR 2016}.

\bibitem[Tang et~al.(2020)Tang, Zhao, Zhang, Gong, Li, and
  Yang]{tang2020integrating}
Yao Tang, Lin Zhao, Shanshan Zhang, Chen Gong, Guangyu Li, and Jian Yang.
\newblock Integrating prediction and reconstruction for anomaly detection.
\newblock \emph{Pattern Recognition Letters}, 129:\penalty0 123--130, 2020.

\bibitem[Zhao et~al.()Zhao, Deng, Shen, Liu, Lu, and Hua]{videoae}
Yiru Zhao, Bing Deng, Chen Shen, Yao Liu, Hongtao Lu, and Xian-Sheng Hua.
\newblock Spatio-temporal autoencoder for video anomaly detection.
\newblock In \emph{ICM2017}.

\bibitem[Subramaniam et~al.()Subramaniam, Nambiar, and Mittal]{cosam}
Arulkumar Subramaniam, Athira Nambiar, and Anurag Mittal.
\newblock Co-segmentation inspired attention networks for video-based person
  re-identification.
\newblock In \emph{ICCV 2019}.

\bibitem[Sosnovik et~al.()Sosnovik, Moskalev, and Smeulders]{smeulders}
Ivan Sosnovik, Artem Moskalev, and Arnold~WM Smeulders.
\newblock Scale equivariance improves siamese tracking.
\newblock In \emph{WACV2021}.

\bibitem[Huang et~al.(2019)Huang, Zhao, and Huang]{got10k}
Lianghua Huang, Xin Zhao, and Kaiqi Huang.
\newblock Got-10k: A large high-diversity benchmark for generic object tracking
  in the wild.
\newblock \emph{IEEE Trans. on Pattern Analysis and Machine Intelligence},
  2019.

\bibitem[Zhang et~al.({\natexlab{c}})Zhang, Wang, Qi, Wang, Feng, and
  Lu]{structsiam}
Yunhua Zhang, Lijun Wang, Jinqing Qi, Dong Wang, Mengyang Feng, and Huchuan Lu.
\newblock Structured siamese network for real-time visual tracking.
\newblock In \emph{ECCV 2018}, {\natexlab{c}}.

\bibitem[Dong and Shen()]{siamtriplet}
Xingping Dong and Jianbing Shen.
\newblock Triplet loss in siamese network for object tracking.
\newblock In \emph{ECCV 2018}.

\bibitem[Li et~al.({\natexlab{b}})Li, Wu, Wang, Zhang, Xing, and
  Yan]{siamrpnPP}
B~Li, W~Wu, Q~Wang, F~Zhang, J~Xing, and J~SiamRPN+ Yan.
\newblock Evolution of siamese visual tracking with very deep networks.
\newblock In \emph{CVPR 2019}, {\natexlab{b}}.

\end{thebibliography}

\end{document}